\documentclass{article} 
\usepackage{colm2024_conference}
\usepackage{graphicx}
\usepackage{microtype}
\usepackage{hyperref}
\usepackage{url}
\definecolor{textblue}{rgb}{.2,.2,.7}
\definecolor{textred}{rgb}{0.54,0,0}
\definecolor{textblack}{rgb}{0,0,0}
\definecolor{textgreen}{rgb}{0,0.43,0}
\usepackage{listings}
\lstdefinestyle{bashstyle}{
    language=bash,
    morekeywords={sudo, apt-get}, 
    breaklines=true,
    commentstyle=\color{textblack},
}

\lstdefinestyle{pythonstyle}{
    language=Python,
    morekeywords={None}, 
    breaklines=true,
}

\usepackage{booktabs}
\usepackage{amsmath}

\title{Octopus: On-device language model for function calling of software APIs}

\author{Wei Chen$^{\dagger}$ \thanks{Corresponding author, $^\dagger$ equal contribution} \\
Stanford University\\
\texttt{\{weichen6\}@stanford.edu} \\
\And
Zhiyuan Li$^\dagger$ \\
Stanford University\\
\texttt{\{zhiyuan8\}@stanford.edu} \\
\AND
Mingyuan Ma$^\dagger$ \\
Harvard University \\
\texttt{\{mingyua\_ma\}@g.havard.edu} \\
}

\colmfinalcopy 
\begin{document}

\maketitle
\begin{abstract}
In the rapidly evolving domain of artificial intelligence, Large Language Models (LLMs) play a crucial role due to their advanced text processing and generation abilities. This study introduces a new strategy aimed at harnessing on-device LLMs for invoking software APIs. We meticulously compile a dataset derived from software API documentation and fine-tune LLMs with capacities of 2B, 3B, and 7B parameters, specifically to enhance their proficiency in software API interactions. Our approach concentrates on refining the models' grasp of API structures and syntax, significantly enhancing the accuracy of API function calls. Additionally, we propose a \textit{conditional masking} technique to ensure outputs in the desired formats and to reduce error rates while maintaining inference speed. We also propose a novel benchmark designed to evaluate the effectiveness of LLMs in API interactions, establishing a foundation for subsequent research. Octopus, the fine-tuned model, is shown to outperform GPT-4 in software API calling. This research aims to advance automated software development and API integration, representing substantial progress in aligning LLM capabilities with the demands of practical software engineering applications.
\end{abstract}

\section{Introduction}

The advent of Large Language Models (LLMs) has revolutionized the field of artificial intelligence, bringing forth a wide array of capabilities in natural language processing, alongside applications in specialized domains such as mathematics (\cite{imani2023mathprompter, he2023solving}), healthcare (\cite{imani2023mathprompter, jo2023understanding, thirunavukarasu2023large}), and legal analysis (\cite{cui2023chatlaw, fei2023lawbench,luppi2022synergistic}). Despite these advancements, LLMs face challenges in assimilating real-time updates and executing specific tasks such as image/video editing (\cite{fu2023guiding}) or intricate tax filings. The integration of LLMs with external APIs emerges as a pivotal improvement. This synthesis, leveraging APIs, not only augments the LLMs' capabilities by facilitating access to up-to-date information and specialized functionalities but also sparks the creation of novel applications such as code interpreters (\cite{bairi2023codeplan,vaithilingam2022expectation,chen2021evaluating}). Research such as ToolAlpaca (\cite{tang2023toolalpaca}) and NexusRaven (\cite{srinivasan2023nexusraven}) also demonstrates the function-calling capability of open-source language models. Consequently, this integration signifies a crucial step toward overcoming the inherent limitations of LLMs, thereby extending their utility and potential for innovation in the field.

Enhancing the integration of Large Language Models (LLMs) with external APIs necessitates addressing the challenge of balancing dependency on large-scale models against efficiency and cost. Many specific tasks utilize only a fraction of the available APIs, which highlights the inefficiency of relying solely on large models such as GPT-4 (\cite{radford2018improving, radford2019language, brown2020language, achiam2023gpt, wu2023empirical}), which require substantial computational resources. This scenario advocates for the development of smaller, task-oriented LLMs that preserve essential functionality while minimizing operational costs (\cite{shen2024hugginggpt, pallagani2024prospects}). However, this shift towards smaller models introduces new challenges, including an increased risk of errors or ``hallucinations'' (\cite{yao2023llm, zhang2023siren, ji2023towards}), which cause issues in precise output formatting (\cite{jiang2023llmparser}); correct output formatting is critical for robust software applications. 

In response to the limitations of oversized Large Language Models (LLMs), which entail unnecessary inference costs and exhibit a lack of focus in training data, we propose a new framework for LLM training and inference. Grounded in an expansive dataset of over 30,000 widely utilized APIs from Rapid API Hub (\cite{rapidapihub}), this framework spans a diverse array of functionalities, from Google searches to Amazon product lookups. By leveraging curriculum learning (\cite{liu2024let}) strategies, we significantly refine the LLMs' proficiency in selecting the appropriate API functions from a pool of similar options. This strategic dataset engineering, combined with our choice of base models, including CodeLlama 7B (\cite{roziere2023code, touvron2023llama}), Google's Gemma 7B \& 2B (\cite{gemma-2023-open-models}), and Stable Code 3B (\cite{stable-code-3b}), underscores the effectiveness of our approach, outperforming GPT-4's benchmarks. Moreover, this ensures the practicality of our solution across various platforms, including mobile devices, since these models can already be deployed on mobile (\cite{mlc-llm}). 

To ensure the consistency of our model's output formatting, we introduce a conditional masking technique during inference. This innovative approach guarantees that our LLMs generate outputs in the desired formats, markedly improving accuracy and reducing validation loss without sacrificing inference speed. We also prove mathematically that the conditional masking technique can only increase accuracy. 

This advancement, validated across our selected base models, not only showcases the potential of compact LLMs in external API integration but also sets a new efficiency benchmark for scalable AI applications. Through a detailed exposition of our model selection and training process, we present a holistic solution that effectively addresses the prevailing challenges in LLM API utility. The dataset used for LLM training and the fine-tuned models will be open-sourced soon.

\section{Related Work}

\textbf{Enhancing LLMs with Tools}\quad  The integration of external computational tools with Large Language Models (LLMs) such as GPT-4, Alpaca, and Llama signifies a substantial advancement in augmenting their capabilities. Initially, integration efforts centered on model-specific fine-tuning methods (\cite{lin2024data,hu2023llm}), which, despite their effectiveness, encountered challenges in widespread and flexible application. A notable shift occurred with the adoption of prompts containing exemplary demonstrations, broadening the scope of tool accessibility. This range includes specialized code interpreters and extensive retrieval frameworks, significantly enhancing the models' ability to interpret and execute complex instructions (\cite{zhou2023llm}). Developments in simulated environments for tool interaction (\cite{shen2024small, du2024anytool, xi2023rise}) and frameworks for API engagement (\cite{li2023api}) have been observed as well. Furthermore, the incorporation of advanced reasoning strategies (\cite{valmeekam2022large, hao2023reasoning, lewkowycz2022solving}) has significantly improved the models' efficiency in interpreting and solving complex tasks.

\textbf{Dataset Format}\quad  The optimization of datasets (\cite{zhuang2024toolqa, kong2023tptu}) for model fine-tuning is critical for enhancing LLM performance. This process involves multi-stage refinement utilizing models such as GPT-4 and Alpaca. By iteratively enhancing the dataset, this methodology not only refines prompts but also improves response quality and develops advanced chain-of-thought (\cite{wang2023chain,zhang2022automatic,shridhar2023art, zheng2023progressive,wei2022chain}) processes. Such advancements lead to a significant increase in the accuracy of function calling within LLMs, setting new benchmarks in dataset optimization and model training. This iterative refinement represents a strategic shift towards enhancing LLM output precision and quality.

\textbf{Robustness in LLM Generation}\quad In contrast to article generation, which may accommodate flexible output formats, software applications necessitate strict adherence to specific output structures, such as the JSON formatting in \cite{zheng2023efficiently}. Many format-consistency problems have been observed in LLM generation (\cite{vaswani2017attention, ackerman2023large}). Some research has been conducted on enforcing these rigid output formats to maintain consistency and reliability in LLM-generated content. For example, the LangChain framework \cite{langchain} provides many output parsers to enforce formats such as \texttt{YAML}, \texttt{JSON}, and \texttt{CSV}. However, there are still many cases that cannot be resolved by output parsers, especially for function-call responses.

\section{Methodology}
In this section, we detail our approach to dataset collection and preparation, introducing the workflow we designed to format the dataset for effective training. We then describe the development of our model, Octopus, highlighting the training techniques and inference strategies we employed. One of the key innovations in our model is the use of a \textit{conditional mask} for inference enhancement, which represents a novel approach to improving model performance. This methodology combines comprehensive data preparation with advanced modeling techniques to address the challenges of training and inference in function-call model development.

\subsection{Dataset collection and refinement}
Our initial dataset comprises API documentation sourced from RapidAPI Hub, one of the world's largest API repositories. This selection was made based on the website's claim of engagement by millions of developers. To facilitate the large language model's comprehension of API usage patterns, we compiled a comprehensive collection of API documentation, focusing on approximately 30,000 of the most frequently utilized APIs. This dataset acquisition was structured in two primary stages: the initial collection and processing of individual API documentation entries, followed by a meticulous refinement process to optimize the dataset for training purposes.

\textbf{Single API Data Preprocessing}\quad Through a detailed exploration of the documentation, we gained a comprehensive understanding of how RapidAPI Hub's API usage examples are structured and utilized. Our approach involves meticulously extracting API usage examples, which detail the API's name, description, argument names, and their respective descriptions, and formatting this information as JSON. This data is then reorganized using the OpenAI GPT-3.5 and CodeLlama 70B models to align with the desired organizational standards. Then, we refine the function names based on their descriptions to ensure they are concise and informative. Subsequently, argument names and descriptions are captured. To counteract potential inaccuracies (``hallucinations'') inherent in smaller LLMs, the Python coding format is employed. This decision is strategic, leveraging the models' inherent code-reasoning capabilities acquired from training on extensive code datasets, as in the CodeLlama 7B and Stable Code 3B models. This process not only streamlines the API information for enhanced usability but also leverages advanced AI models to ensure the information is presented in a structured, accessible manner. By prioritizing the function description as a guide for renaming and carefully detailing argument names and descriptions, the approach ensures that the essential elements of API usage are conveyed effectively, supporting developers in integrating these APIs into their projects seamlessly. An example of a converted function is shown below.

\begin{lstlisting}[style=pythonstyle]
def get_flight_details(flight_id):
  """
  Get detailed information on specific flights, including real-time tracking,  departure/arrival times, flight path, and status insights.
  Args:
    flight_id (string): The flight_id represents the ID of a flight.
  """
\end{lstlisting}

In our methodology, we deliberately excluded function bodies from the final dataset compilation. Through a meticulous selection process, we aggregated approximately 20,000 APIs, employing OpenAI GPT-4 for a comprehensive examination and the removal of APIs with deficiencies, such as missing arguments or inconsistencies between function descriptions and their parameters. This stringent selection criterion was pivotal in assuring the dataset's quality. Each API underwent this rigorous scrutiny, culminating in the compilation of dataset A, which serves as the basis for the subsequent data processing.

\begin{figure}[h]
    \centering
    \includegraphics[width=0.8\textwidth]{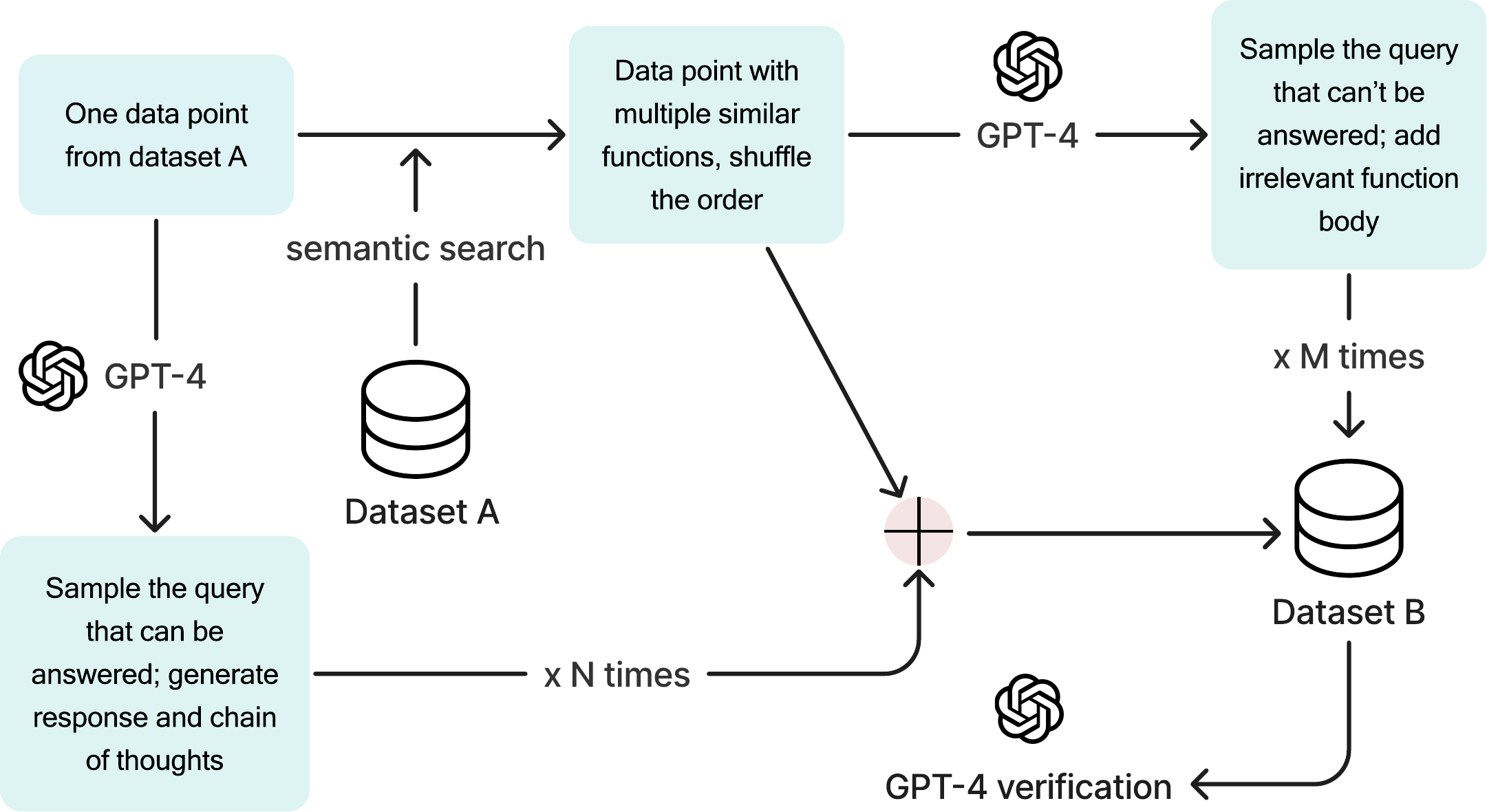}
    \caption{Refining dataset A into dataset B through a strict workflow. This process involves three critical steps: sampling positive queries solvable by specific APIs and generating the corresponding responses and CoTs; identifying unsolvable queries and augmenting them with irrelevant function bodies; and employing semantic analysis to incorporate similar functions into data points. Following GPT-4's rigorous verification, dataset B emerges as the optimized training dataset, poised to significantly elevate model efficacy.}
    \label{fig:ds-pipeline}
\end{figure}

\textbf{Dataset Refinement}\quad To enhance decision-making in Large Language Models (LLMs) for real-world API usage, we present a sophisticated dataset construction approach that is crucial to our study. We begin by integrating various functions, intentionally incorporating some irrelevant functions to create a complex environment for the LLM. Inspired by curriculum learning, we design our dataset to include hard negative samples gradually. This involves introducing similar functions to incrementally raise the difficulty of selecting the most relevant function. Our approach is depicted in Figure (\ref{fig:ds-pipeline}), illustrating the detailed process of compiling the dataset. Below, we describe the techniques employed.

\begin{enumerate}
    \itemsep0em
    \parsep0em
    \item \textbf{Negative samples}\quad To enhance the model's reasoning capabilities and practical applicability, our methodology involves sampling both positive and negative examples. The ratio of these datasets is represented by the variable $\frac{M}{N}$ in Figure (\ref{fig:ds-pipeline}), serving as an important parameter in our experimental setup. Specifically, in our framework, we select $M$ and $N$ to be equal, setting both values to 1. 
    \item \textbf{Similar function clusters}\quad In our practical implementation, the model selects functions from a diverse pool in response to user queries. To intensify the training challenge, we deliberately complicate the selection process. Specifically, we construct training data by associating a given data point with three semantically similar ones. This process involves calculating vector embeddings from function descriptions, with Milvus facilitating the search. The sampling of the three similar functions is determined by their similarity scores, focusing on ranks 5 to 10, to deliberately exclude overly similar functions and avoid redundancy in individual queries. This approach guarantees a challenging training setting, cultivating a model capable of differentiating between closely related functions in practical use cases.
    \item \textbf{GPT-4-generated queries}\quad The creation of a high-quality dataset depends crucially on the formulation of qualified queries. In this context, we opt to generate positive queries solvable by a single API. Moreover, for such positive instances, we also generate and incorporate a Chain of Thought (CoT), which is utilized during model training. Recent studies have demonstrated that the addition of CoT not only enhances model performance but also significantly improves its reasoning abilities (\cite{srinivasan2023nexusraven}). Notably, the creation of qualified queries and auxiliary information is crucial to developing effective training datasets.
    \item \textbf{GPT-4 verification}\quad During our dataset's development, we observed that GPT-4-generated responses can contain errors, despite the model's advanced capabilities. Thus, we designed a workflow that lets GPT-4 conduct self-verification, effectively identifying and rectifying inaccuracies in its outputs. After obtaining dataset B, we employed GPT-4 to meticulously verify and eliminate any data points that failed to meet our stringent quality criteria. This rigorous validation process led to the exclusion of approximately 1,000 data points, significantly contributing to our model's enhanced performance.
\end{enumerate}

Adhering to the outlined methodology, we meticulously compiled the training dataset, achieving an impressive collection of approximately 150,000 data points. Each individual API is associated with 5 positive queries that it can resolve. To provide a comprehensive understanding, a sample from the complete dataset is included in the Appendix (\ref{ds:eg1}), showcasing the detailed structure and composition of our training data.

\subsection{Octopus}
To validate the efficacy of our framework, we fine-tuned four renowned open-source models: CodeLlama 7B, Google Gemma 2B \& 7B, and Stable Code LM 3B. A standardized training template, detailed in the Appendix (\ref{ds:eg1}), was employed across all models. We utilized LoRA and 8-bit quantization techniques, allocating A100 80GB GPU hours as follows: 90h for CodeLlama 7B and Google Gemma 7B, 30h for Google Gemma 2B, and 60h for Stable Code LM 3B. The learning rate was set to 5$\times$10$^{-5}$, with a linear scheduler optimizing outcomes. In the inference stage, user queries trigger function retrieval and execution, mapping the generated function and its arguments to the corresponding APIs for the final responses, thus ensuring accurate results upon correct generation of the function and argument names.

\begin{figure}[h]
    \centering
    \includegraphics[width=1.0\textwidth]{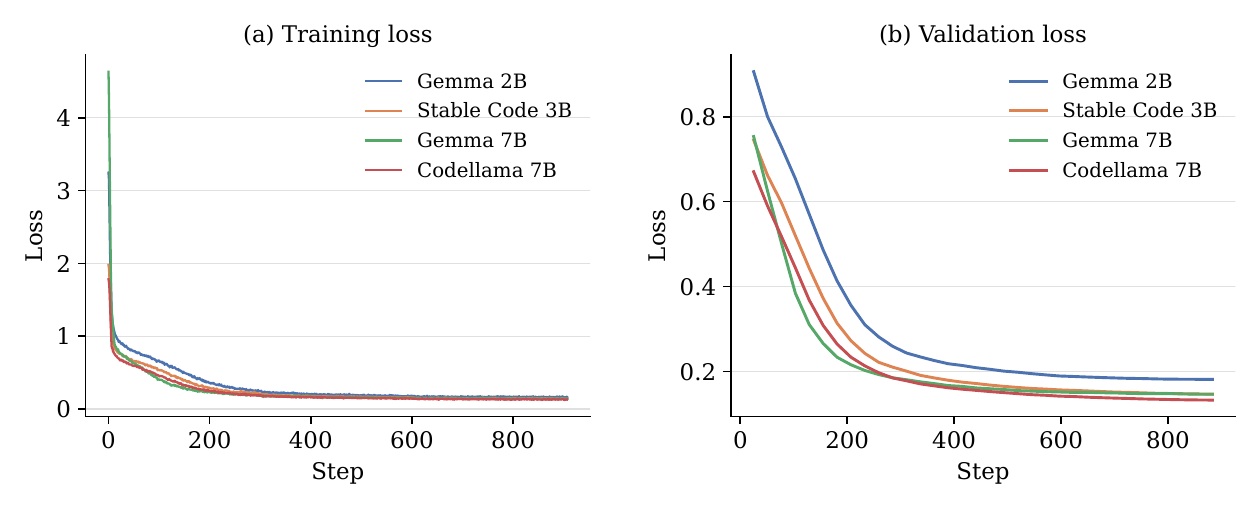}
    \caption{The training and validation loss for selected pretrained models.}
    \label{fig:loss}
\end{figure}

We experimented with different LoRA setups and found that the best configuration is to choose a LoRA rank of 16 and apply the method to the modules \texttt{"q\_proj", "v\_proj", "o\_proj", "up\_proj", "down\_proj"}. We also present the training and validation loss for selected models in Figure (\ref{fig:loss}). During training, we progressively trained on data points with more similar examples to implement curriculum learning.

\subsection{Inference using conditional mask}
The utilization of smaller-parameter Large Language Models (LLMs) poses a pivotal challenge: a noticeable decrement in robustness when generating outputs. This challenge is also observed in our model, which necessitates enforcing responses with precise function names along with their corresponding arguments. The expected output format demands that arguments be encapsulated within parentheses, that function names align with a pre-defined repository, and that argument values conform to their designated types. Discrepancies, such as typographical errors in function names or misalignment in argument types, critically undermine the integrity of the output, rendering it susceptible to errors. For instance, in both GPT-4 and our model, deviations in the function name—whether through misspelling or elongated expressions—can lead to unintended corrections that fail to map back to the original function names, thereby distorting the intended output. The original LLM generation process samples the next token as 
\begin{equation}
    P(x_{t+1}|x_{1:t}) = P(x_{t+1}|x_{1:t}; \operatorname{LLM}),\quad x_{t+1}=\operatorname{argmax} P(x_{t+1}|x_{1:t}; \operatorname{LLM})  
\end{equation}
where $x_{1:t}$ denotes all the current tokens, with sequence length $t$, and $x_{t+1}$ is the next token to be sampled. What we do here is introduce an additional mask dependent on $x_{1:t}$, so that 
\begin{equation}
    x_{t+1}=\operatorname{argmax} \left[P(x_{t+1}|x_{1:t}; \operatorname{LLM})\odot \operatorname{mask}(x_{1:t})  \right]. 
\end{equation}

In constructing the mask, we designate all tokens that do not align with the correct format to be masked by assigning a value of 0 to their respective positions, and a value of 1 to all other positions. For example, if we already know that the next token represents an integer, we only unmask the tokens that are used for integers. Therefore, the formulation of an accurate \textit{mask} is paramount for achieving the desired outcome. In this context, we delineate several methodologies that were investigated for the derivation of the \textit{mask}.

\begin{itemize}
    \item \textbf{Enum data type}\quad Function names are usually known in advance and do not change during inference, so we can treat them as enumerable data variables. To efficiently manage these names, a Trie can be constructed, facilitating retrieval of the \textit{mask} with a time complexity of $O(D)$, where $D$ denotes the Trie's depth, equivalent to the maximum length of a function name, which in our case is approximately 20. This results in constant time complexity. As an alternative approach, storing all prefixes of potential function names within a dictionary could further reduce the complexity to $O(1)$. The implementation of the Trie class is provided in the Appendix (\ref{ds:eg2}).
    \item \textbf{String, float, dict, and int types}\quad Regular expressions can be employed to analyze subsequent tokens and generate the conditional mask.
\end{itemize}

Therefore, we can ensure that the output is free from formatting errors.
Our experimental findings indicate that the application of the conditional mask significantly enhances the robustness of the Large Language Model (LLM) in the context of function calls.

\section{LLM Evaluation for Function Calling}

We conducted a comprehensive series of tests on our dataset, evaluating the Octopus model against other leading models. This evaluation focused on Octopus's capability to understand API calls, specifically those on RapidAPI. Additionally, we explored the impact of incorporating various retrieval techniques during the training phase on the model's ultimate effectiveness.

In terms of baselines, our primary comparison was with cutting-edge language models in a zero-shot framework. The models we analyzed include GPT-4 by OpenAI, utilizing the gpt-4-0314 checkpoint, and GPT-3.5-turbo, employing the gpt-3.5-turbo-0301 checkpoint. Both models have been refined through Reinforcement Learning from Human Feedback (RLHF) for enhanced conversational abilities. 

\subsection{Evaluation dataset and benchmark}
To benchmark function calls within commonly used software APIs, we constructed a dedicated dataset. This dataset was generated by randomly selecting four different function APIs and sampling queries that could be addressed by these APIs. The sampling utilized the same prompt template employed during training, details of which are provided in Appendix \ref{ds:eg1}. Additionally, we included queries that these APIs could not resolve, maintaining a balanced ratio of solvable to unsolvable queries of 1:1. Ground truth for the dataset was established through \textit{human annotation}. We applied rigorous standards for benchmarking, focusing on real-world application requirements, including the precise matching of function names and arguments. For models not trained on this dataset, issues related to format correctness were overlooked to provide a fairer comparison. Consequently, in our analysis, GPT-3.5 was not marked incorrect for format discrepancies.

\begin{table}[ht]
\centering
\caption{Function-call accuracy of GPT-3.5, GPT-4, and the fine-tuned ``Octopus'' series models, evaluated with and without the conditional mask. The prefix ``Octopus'' denotes the series, while the suffix indicates the specific pretrained base model. The conditional mask cannot be applied to GPT-3.5 and GPT-4, whose APIs do not expose logits.}
\label{tab:accuracy}
\vspace{4pt}
\setlength{\tabcolsep}{10pt}
\begin{tabular}{@{}lcc@{}}
\toprule
& \multicolumn{2}{c}{Accuracy (\%)} \\
\cmidrule(l){2-3}
Model & Without mask & With mask \\
\midrule
GPT-3.5              & 50 & --- \\
GPT-4                & \textbf{96} & --- \\
\midrule
Octopus-codellama7B  & 92 & \textbf{97} \\
Octopus-gemma7B      & 93 & \textbf{97} \\
Octopus-stablecode3B & 93 & 95 \\
Octopus-gemma2B      & 93 & 94 \\
\bottomrule
\end{tabular}
\end{table}

\subsection{Performance without the conditional mask}
For the task of inferring function calls, we initially employed both the GPT-3.5 and GPT-4 models to generate responses. For these pretrained models, greedy search was utilized to select responses. This decision was based on the higher precision required for accurately identifying both function names and their corresponding parameters, where the model's ability to choose the correct function name and parameters is crucial. Therefore, alternative methods such as sampling and beam search were not adopted for this task. The resulting accuracy metrics from this approach are presented in Table~\ref{tab:accuracy}.

The results highlight that GPT-4 consistently achieves the highest accuracy in producing correct outcomes. A notable issue leading to inaccuracies with GPT-4 involves ``hallucinations,'' such as its tendency to autocorrect misspelled function names, exemplified by transforming \texttt{send\_emil} into \texttt{send\_email}. It is critical that the function name provided in the initial prompt remain unaltered, regardless of spelling errors. This correction issue extends to parameters as well; for instance, GPT-4 might generate \texttt{Australian} as a parameter when the query explicitly requires a country name. The primary source of incorrect outputs is attributed to the generation of inaccurate parameters. However, all pretrained models demonstrate near-perfect performance in identifying the correct function name.

\subsection{Performance with the conditional mask}
In contrast to the inference approach described in the preceding subsection, we implemented a conditional mask during inference for this scenario. This modification has been effective in enhancing outcomes, particularly in the generation of parameters. Utilizing a conditional mask, especially when an input is of an enum type such as a country name, helps prevent the model from generating unexpected parameters. The improvements facilitated by this approach are shown in Table~\ref{tab:accuracy}. However, since the APIs for GPT-3.5 and GPT-4 do not provide logits, the conditional masking technique could not be applied to these models, and thus no improvement in their metrics was observed. Nevertheless, it is noteworthy that the two 7B models were able to achieve performance surpassing GPT-4, highlighting the efficacy of the conditional masking technique in certain contexts.

\section{Conclusion}
In this study, we present a novel framework designed to train large language models on practical software APIs, together with a subsequent evaluation of their performance in making API calls, specifically in comparison to the GPT-4 model. Our approach includes a methodology for refining the dataset and the associated prompt template, incorporating negative sampling and curriculum learning strategies. Additionally, we introduce an innovative technique known as the conditional mask, aimed at addressing the challenge of mismatched output formats.

\subsubsection*{Acknowledgments}
We acknowledge the significant contributions of the teams at Google, CodeLLAMA, and Stable AI in advancing the open model ecosystem through their provision of powerful pretrained models. These contributions have been instrumental in the development of Octopus.

\bibliography{colm2024_conference}
\bibliographystyle{colm2024_conference}

\appendix
\section{Mathematical Derivation}
\subsection{Impact of conditional masking on inference performance}

In this appendix, we examine the effect of applying a conditional mask during inference on a causal language model's accuracy and validation loss. Consider the validation loss without masking defined as:

\begin{equation}
    L_{\text{val}}^{\text{non-mask}} = \sum_{i \in V} -y_i \log(\hat{y}_i),
\end{equation}
where $V$ denotes the vocabulary set, and $y_i$ is a binary indicator (0 or 1) of whether class label $i$ is the correct classification for the current observation. 

Introducing a conditional mask allows us to partition the vocabulary $V$ into two subsets: $V_1$, containing the indices that are not masked, and $V_2$, containing the indices that are masked. Given that the true label $y_i$ belongs to $V_1$ during inference, and considering that for all $i$,

\begin{equation}
    -y_i\log(\hat{y}_i) > 0,
\end{equation}

the validation loss with masking can be expressed as:

\begin{equation}
    L_{\text{val}}^{\text{mask}} = \sum_{i \in V_1} -y_i \log(\hat{y}_i) < L_{\text{val}}^{\text{non-mask}},
\end{equation}

indicating that the validation loss is reduced when a conditional mask is applied during inference.

Accuracy, particularly precision in this context, for the non-masked scenario is determined by the alignment between the ground truth label's index and the index of the maximum value in the predicted distribution:

\begin{equation}
    \text{Precision}^{\text{non-mask}} = \mathbb{1}[\text{argmax}_i (y_i) = \text{argmax}_i (\hat{y}_i)],
\end{equation}
where $\mathbb{1}[\cdot]$ is the indicator function, returning 1 if the condition is true, and 0 otherwise.

With conditional masking, the prediction $\hat{y}_i$ is constrained to $V_1$, effectively reducing the search space for $\text{argmax}_i (\hat{y}_i)$ and increasing the likelihood of matching $\text{argmax}_i (y_i)$, given that $y_i \in V_1$. Hence,

\begin{equation}
    \text{Precision}^{\text{mask}} \geq \text{Precision}^{\text{non-mask}},
\end{equation}

demonstrating that conditional masking during inference not only reduces validation loss but also enhances the model's precision by focusing on a more relevant subset of the vocabulary.

\section{Dataset and code illustration}\label{ds:eg}
\subsection{Dataset template}\label{ds:eg1}
\begin{lstlisting}[style=pythonstyle]
"""
You are an assistant, and you need to call find appropriate functions according to the query of the users. Firstly, find the relevant functions, then get the function arguments by understanding the user's query. The following functions are available for you to fetch further data to answer user questions:

Function:

def no_relevant_function(user_query):
  '''
  Call this when no other provided function can be called to answer the user query.
  Args:
    user_query (str): The user_query that cannot be answered by any other function calls.
  '''


def youtube_downloader(videourl):
  '''
  Get direct video URL for youtube to download and save for offline viewing or sharing.
  Args:
    videourl (string): The URL of the video being accessed as a string.
  '''


def facebook_dl_link(url):
  '''
  Get downloadable link for facebook, allowing convenient offline viewing and sharing.
  Args:
    url (string): The URL string for the function argument.
  '''


def pinterest_video_dl_api(url):
  '''
  Get download feature for videos from Pinterest enabling users to save videos for offline viewing.
  Args:
    url (string): The URL string represents the web address of the resource being accessed.
  '''


def insta_download_url(url):
  '''
  Get download access to Instagram content by inputting the URL, enabling users to save and view content offline.
  Args:
    url (string): The URL string.
  '''


Obtain download access for viewing a recent Instagram post offline using the URL https://www.instagram.com/p/CODEinstantiate123/

Response:insta_download_url('https://www.instagram.com/p/CODEinstantiate123/')<nexa_end>

Thought:To acquire download access for Instagram content for offline viewing, 'insta_download_url' is called with the post's URL as the argument, ensuring the content specified by the URL is fetched for download.    
"""
\end{lstlisting}

\subsection{Trie class to process the enum variable}\label{ds:eg2}
\begin{lstlisting}[style=pythonstyle]
class TrieNode:
    def __init__(self) -> None:
        self.children: Dict[str, TrieNode] = {}
        self.isEndOfWord: bool = False


class Trie:
    def __init__(self) -> None:
        self.root: TrieNode = TrieNode()
    
    def insert(self, word: str) -> None:
        node = self.root
        for char in word:
            if char not in node.children:
                node.children[char] = TrieNode()
            node = node.children[char]
        node.isEndOfWord = True
    
    def is_prefix(self, prefix: str) -> bool:
        node = self.root
        for char in prefix:
            if char not in node.children:
                return False
            node = node.children[char]
        return True
    
    def get_all_prefixes(self) -> List[str]:
        prefixes: List[str] = []
        self._dfs(self.root, "", prefixes)
        return prefixes
    
    def _dfs(self, node: TrieNode, prefix: str, prefixes: List[str]) -> None:
        if node != self.root:
            prefixes.append(prefix)
        for char, next_node in node.children.items():
            self._dfs(next_node, prefix + char, prefixes)
    
    def search(self, prefix: str, include_prefix: bool = True) -> List[str]:
        node = self.root
        for char in prefix:
            if char not in node.children:
                return []
            node = node.children[char]
    
        initial_string: str = prefix if include_prefix else ""
        return self._find_words_from_node(node, initial_string)
    
    def _find_words_from_node(self, node: TrieNode, current_string: str) -> List[str]:
        words: List[str] = []
        if node.isEndOfWord:
            words.append(current_string)
        for char, next_node in node.children.items():
            words.extend(self._find_words_from_node(next_node, current_string + char))
        return words
\end{lstlisting}

\end{document}